\documentclass{article}
\usepackage[final,main]{neurips_2026}
\setcitestyle{numbers,square,comma}
\usepackage[utf8]{inputenc}
\usepackage[T1]{fontenc}
\usepackage{url}
\usepackage{booktabs}
\usepackage{amsfonts}
\usepackage{nicefrac}
\usepackage{xcolor}
\usepackage{amsmath,amssymb}
\usepackage{graphicx}
\usepackage{float}
\usepackage{flafter}
\usepackage{placeins}
\usepackage{makecell}
\usepackage{multirow}
\usepackage{hyperref}
\hypersetup{pdftitle={The Override Gap: A Magnitude Account of Knowledge Conflict Failure in Hypernetwork-Based Instant LLM Adaptation},pdfauthor={Shuaizhi Cheng, Xiang Shi, Zhiwei Zhang, Mingwei Li}}

\newcommand{\prediction}[2]{\noindent\textbf{#1:} #2\par\vspace{0.05em}}

\title{The Override Gap: A Magnitude Account of Knowledge Conflict Failure\\in Hypernetwork-Based Instant LLM Adaptation}

\author{
Shuaizhi Cheng$^{1,2}$ \quad Xiang Shi$^{2}$ \quad Zhiwei Zhang$^{2,3}$ \quad Mingwei Li$^{2}$\\
$^1$Harbin Institute of Technology \quad $^2$KigLand Machine Learning Lab\\
$^3$CEREA, ENPC, EDF R\&D, Institut Polytechnique de Paris, \^{I}le-de-France\\
{\tt\small szcheng@stu.hit.edu.cn}\quad {\tt\small kevin@kig.land}\\
{\tt\small zhiwei.zhang@enpc.fr}\quad {\tt\small remi@kig.land}
}
\date{}

\begin{document}
\maketitle

\begin{abstract}
Hypernetwork-based methods such as Doc-to-LoRA internalize a document into an LLM's weights in a single forward pass, but struggle when the document contradicts pretraining knowledge. We study this override gap through the competition between the pretrained answer margin and the shift induced by the generated adapter. On 194 Gemma conflicts, baseline accuracy falls from 65.3\% in the weakest-prior quartile to 16.3\% in the strongest. We introduce two training-free methods: Selective Layer Boosting (SLB), which scales the adapter at its highest-activity layers, and Conflict-Aware Internalization (CA), which chooses between standard and stronger adaptation using a short base-model probe. Strong boosting raises Gemma deep-conflict accuracy from 46.4\% to 71.0\%, with conflict gains also observed on Mistral-7B and Qwen-4B. Matched-norm controls show that both update magnitude and allocation matter: SLB benefits from the generated layer and direction structure, while global scaling is competitive at the same increment budget. CA offers a workload-dependent recall--latency tradeoff, reaching 98.4\% novel recall on Gemma while retaining the strong path's conflict accuracy. KID-Bench provides 486 questions spanning novel recall, mixed knowledge, and prior-graded conflicts, with answer-level semantic evaluation.
\end{abstract}

\section{Introduction}

\begin{figure}[htbp]
\centering
\includegraphics[width=\textwidth]{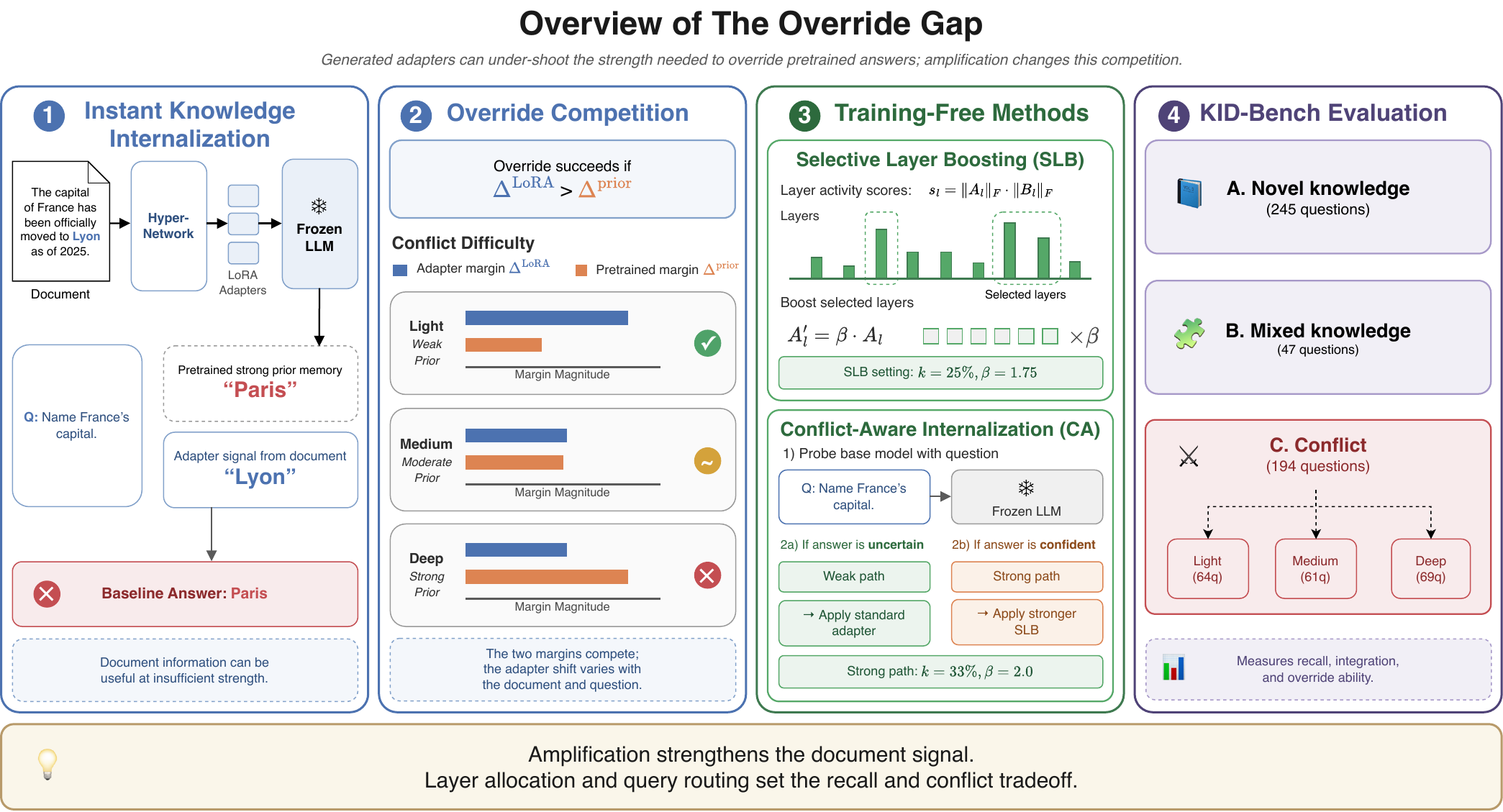}
\caption{Overview of the override gap. A generated adapter changes the competition between document and pretrained answers. SLB amplifies high-activity layers; CA selects the adaptation strength using a base-model probe. The margin bars illustrate the competition schematically.}
\label{fig:overview}
\end{figure}

Hypernetwork-based instant knowledge internalization generates a small LoRA adapter from a document in one forward pass, incorporating document knowledge into model parameters without deployment-time gradient updates~\cite{doc2lora, shine}. It sits between retrieval-augmented generation~\cite{rag, atlas}, which keeps the document in context, and gradient fine-tuning~\cite{lora}, which takes minutes to hours. Existing evaluations mostly test recall of stated facts, but rarely examine whether the model follows the document when it \emph{contradicts} knowledge the model already holds, e.g., changed regulations, product updates, or primary-source corrections~\cite{kc_survey, adaptive_chameleon}.

To evaluate whether models follow documents that contradict pretrained knowledge, we introduce KID-Bench\footnote{Code and data: \url{https://github.com/KigML/OverrideGap}.}, covering novel recall, mixed knowledge, and conflicts of varying prior strength. We find a substantial override gap: the Gemma baseline succeeds on 57.8\% of light conflicts, 50.8\% of medium conflicts, and 46.4\% of deep conflicts. A magnitude account connects this failure to the relative strength of document and pretrained answer signals. The generated adapter changes the relevant answer margin, but its strength is not explicitly calibrated to the prior being contradicted. When the adapter carries a useful directional signal, amplification can make the document answer preferred and narrow the override gap.

This paper advances three contributions. \textbf{(i)} We formalize the pairwise override condition as $\Delta^{\text{lora}}>\Delta^{\text{prior}}$ and study its relationship to measured prior strength and scaling interventions. \textbf{(ii)} We introduce Selective Layer Boosting (SLB) and Conflict-Aware Internalization (CA), then separate their contributions through matched-norm layer and direction controls, alongside an always-strong routing baseline. Strong boosting raises Gemma deep-conflict accuracy from 46.4\% to 71.0\%; CA provides an optional recall--latency tradeoff. \textbf{(iii)} We introduce KID-Bench, a 486-question benchmark for novel recall, mixed knowledge, and prior-graded conflicts. Semantic scoring and document-paired comparisons connect the benchmark's measurements to the behavior each question asks for.

\section{Background}

Document-based adaptation incorporates new information into a model's parameters. Transformer feed-forward networks~\cite{transformer} can act as key-value memories for factual associations~\cite{ff_keyvalue, knowledge_neurons, rome, dissecting_recall, superposition}, making them a natural target for factual updates. LoRA~\cite{lora} provides a parameter-efficient form of these updates by factorizing a weight update as $\widetilde{W} = W + (\alpha/r) B A$ with $A \in \mathbb{R}^{r \times d'}$ and $B \in \mathbb{R}^{d \times r}$, where $r$ is the adapter rank, $\alpha$ is the scaling factor, and $d',d$ are the input and output dimensions. The pretrained $W$ is frozen; only $A, B$ are trained. 

Instant knowledge internalization generates these low-rank factors directly from a document using a hypernetwork~\cite{hypernetworks}, a network that outputs the parameters of another network. Given a document $d$, the hypernetwork $H_\theta$ produces per-layer adapters $\{A_l, B_l\}_{l=1}^{L} = H_\theta(d)$ in a single forward pass. Doc-to-LoRA~\cite{doc2lora} implements $H_\theta$ as a Perceiver-style cross-attention module~\cite{perceiver} that outputs rank-8 adapters for the feed-forward down-projection layers. SHINE~\cite{shine} reuses the backbone LLM as its own encoder. The hypernetwork is trained to reconstruct document content. 

\section{Related Work}

Updating a pretrained model with new information is a central goal of \textbf{knowledge editing}. ROME~\cite{rome}, MEMIT~\cite{memit}, MEND~\cite{mend}, SERAC~\cite{serac}, and Transformer-Patcher~\cite{transformer_patcher} pursue targeted factual changes through parameter updates or additional response mechanisms, with possible effects on related facts~\cite{rippleedits}. When the information is supplied as a document rather than an explicit factual edit, the task becomes incorporating its contents into the model. \textbf{Instant knowledge internalization} addresses this setting through generated adapters, as in Doc-to-LoRA~\cite{doc2lora} and SHINE~\cite{shine}. Related uses of LoRA include task adaptation from language descriptions in Text-to-LoRA~\cite{text2lora}, visual capability learning in Vision as LoRA~\cite{vora}, and hypernetwork-based adapter merging for subject--style image generation in LoRA.rar~\cite{lorarar}. For document internalization, recalling content is only part of the task: when a document contradicts pretrained knowledge, the model must also favor its answer over the existing one.

This competition is studied in work on \textbf{knowledge conflicts}, including ConflictBank~\cite{conflictbank}, adaptive-versus-stubborn model behavior~\cite{adaptive_chameleon}, and broader surveys~\cite{kc_survey}. For generated adapters, it raises a specific question: if an adapter supports the document answer but fails to override the pretrained answer, can increasing its update magnitude help? Addressing this question requires controlling how strongly the adapter is applied. Related mechanisms appear in \textbf{modulation and gating}: FiLM~\cite{film} modulates features, while adaptive LoRA methods~\cite{mole, xlora, adalora, dora, hyperlora} explore different ways to control adaptation. Our work examines whether training-free, post-hoc control of adapter update magnitude can improve conflict resolution in hypernetwork-based instant knowledge internalization.

\section{A Magnitude Account of Conflict Failure}
\label{sec:theory}

An adapter can shift the model's preference toward the document answer without making it preferred over the pretrained answer. To characterize this distinction, we compare the pretrained answer margin with the change induced by the adapter.

\subsection{The Override Competition}

For a question $x$, consider two competing answer tokens at the same answer position: $y_{\text{doc}}$ corresponding to the document answer and $y_{\text{pre}}$ corresponding to the pretrained answer. Let $\ell^{\text{pre}}(y;x)$ denote the base model's logit for token $y$ at this position. We define the \emph{pretrained margin} as
\begin{equation}
\Delta^{\text{prior}}:=\ell^{\text{pre}}(y_{\text{pre}};x)-\ell^{\text{pre}}(y_{\text{doc}};x).
\label{eq:prior_def}
\end{equation}
A positive margin indicates a preference for the pretrained answer; larger positive values indicate a stronger preference. Applying the LoRA adapter adds $(\alpha/r)B_l A_l h_l^{\text{in}}$ to the output of the adapted down-projection at layer $l$, where $A_l$ and $B_l$ are the LoRA factors at layer $l$, $h_l^{\text{in}}$ is the projection's input vector, $r$ is the adapter rank, and $\alpha$ is the LoRA scaling hyperparameter.

Let $\delta\ell(y)$ denote the change in the logit for token $y$ after applying the adapter. This change includes effects propagated through subsequent layers. For simplicity, we omit the dependence on $x$. The adapter's net contribution is
\begin{equation}
\Delta^{\text{lora}}:=\delta\ell(y_{\text{doc}})-\delta\ell(y_{\text{pre}}).
\label{eq:lora_margin}
\end{equation}
The adapted model prefers the document token exactly when
\begin{equation}
\ell^{\text{pre}}(y_{\text{doc}})+\delta\ell(y_{\text{doc}})
>\ell^{\text{pre}}(y_{\text{pre}})+\delta\ell(y_{\text{pre}}),
\end{equation}
which gives the override condition
\begin{equation}
\boxed{\Delta^{\text{lora}}>\Delta^{\text{prior}}}.
\label{eq:compete}
\end{equation}

For $0<\Delta^{\text{lora}}\leq\Delta^{\text{prior}}$, the adapter shifts the model's preference toward the document answer but does not make it preferred over the pretrained answer. Amplification aims to address this limitation. Scaling $A_l$ by $\beta$ scales that layer's weight update exactly; the resulting logit response is locally approximated by a first-order expansion and can become nonlinear across layers. For multi-token answers, Eq.~\eqref{eq:compete} describes the preference between two tokens at a given answer position, while free-generation accuracy evaluates the correctness of the full answer.

\subsection{Adapter Strength and Empirical Predictions}

The hypernetwork's document reconstruction objective does not explicitly condition its output on the pretrained margin of a future query. The generated adapter strength can therefore vary independently of the prior it must overcome. Training frequency motivates one possible source of prior strength~\cite{superposition}; we measure the model's answer probabilities directly. This account motivates four empirical predictions:

\prediction{Prior strength}{Conflict failure should be associated with stronger pretrained answer preferences.}
\prediction{Update magnitude}{Increasing adapter magnitude should improve override where the generated direction is useful, with eventual saturation or interference.}
\prediction{Layer and direction allocation}{At a fixed update budget, the allocation of amplification across layers and directions should affect conflict resolution and retention.}
\prediction{Query routing}{Routing can preserve recall when a weak branch answers correctly and a strong branch disrupts that answer; the benefit depends on identifying these cases.}

We examine these predictions through per-question priors, scaling sweeps, matched-norm controls, and paired weak/strong outputs.

\section{Methods}

The magnitude account leads to two complementary choices: where to amplify an update and when to apply it more strongly. SLB uses layer activity for the first choice, adding layer ranking and matrix scaling; CA uses a short base-model probe for the second.

\subsection{Selective Layer Boosting (SLB)}

SLB scales the generated adapter at layers selected by the activity score $s_l=\|A_l\|_F\|B_l\|_F$. Let $k$ denote the selected layer fraction and $L$ the number of adapted layers. We select the $\lfloor kL\rfloor$ layers with the highest activity scores, denoted by $S$, and scale their $A$ matrices:
\begin{equation}
A_l'=\begin{cases}\beta A_l & l\in S,\\ A_l & \text{otherwise}.\end{cases}
\end{equation}
We use $k=25\%$, $\beta=1.75$ for Gemma and Mistral, selected from the layer-fraction and strength sweeps (Appendices~\ref{app:full_ablation} and~\ref{app:beta_sweep}). Qwen uses $\beta=3.0$, a coarse calibration motivated by its smaller adapter norms and checked empirically. Selection is recomputed for each document. The product-of-norms score is a ranking heuristic; control budgets use the effective update norm $\| (\alpha/r)B_lA_l\|_F$.

\subsection{Conflict-Aware Internalization (CA)}

CA first queries the base model without an adapter, generating a probe answer of up to 20 tokens. If the probe contains an uncertainty expression such as ``don't know'' or ``not sure,'' CA uses the standard adapter ($\beta=1$). Otherwise, it applies stronger SLB to approximately the top third of layers ranked by the activity score, using $\beta=2$ for Gemma/Mistral or $\beta=3$ for Qwen. This rule uses the base model's expressed uncertainty to select the adaptation strength. To assess the benefit of this selection, we compare CA with applying the same strong branch to every query. Appendix~\ref{app:ca_impl} provides the uncertainty markers and alternative routing policies.

\paragraph{Relevance Gate (RG-CA).} For queries unrelated to the document, a complementary relevance gate bypasses the adapter. This separates the decision to use document knowledge from CA's choice of adaptation strength. The gate uses alphabetic content-token overlap, a stopword filter, and optional acronym expansion (Appendix~\ref{app:rg_details}).

\section{Evaluation}
\label{sec:kidbench}

We introduce KID-Bench, a benchmark dataset to evaluate whether document-based adaptation enables models to recall new facts, use document facts alongside general knowledge, and follow documents that contradict pretrained knowledge. It contains 486 questions across three categories. \textbf{A (novel, 245 q)} asks about fictional entities described in the document. \textbf{B (mixed knowledge, 47 q)} covers general-knowledge questions and questions that combine document facts with general knowledge. \textbf{C (conflict, 194 q)} contradicts pretrained facts at three qualitative levels: C-Light (64 q), C-Medium (61 q), and C-Deep (69 q). These levels describe the intended strength of the contradicted facts; per-question answer probabilities provide a model-specific measurement. Questions include multiple phrasings of the same knowledge point.

\paragraph{Scoring.} We score the answer's meaning against the document and question, accepting valid aliases and rejecting responses that quote an answer only to deny it. All methods are evaluated on the same 486 questions. For unresolved answer judgments, we report the accuracy range obtained by treating those labels as incorrect or correct. These ranges describe label uncertainty; sampling uncertainty is reported separately using paired bootstrap intervals. Appendix~\ref{app:scoring} gives the scoring criteria, full results, and a supplementary compositional QA evaluation.

\paragraph{Setup.} We use released Doc-to-LoRA~\cite{doc2lora} hypernetwork checkpoints for Gemma-2-2B-IT~\cite{gemma} (80K steps), Mistral-7B-Instruct-v0.2~\cite{mistral7b} (20K steps), and Qwen3-4B-Instruct-2507~\cite{qwen} (20K steps), with greedy decoding and a maximum of 64 generated tokens per answer. Standard QA uses a fixed 200-question SQuAD~\cite{squad} subset. All interventions reuse the same backbone and generated adapter within a comparison. The baseline applies the generated adapter without amplification; always-strong applies CA's strong branch to every query. We include retrieval-augmented generation (RAG)~\cite{rag, atlas} as a reference baseline under the same no-gradient-at-deployment constraint.

\paragraph{Matched comparisons.} SLB and global scaling at the same $\beta$ apply different total increments. We match the increments using $U_l=(\alpha/r)B_lA_l$ and the selected energy fraction
\begin{equation}
f=\frac{\sum_{l\in S}\|U_l\|_F^2}{\sum_l\|U_l\|_F^2},\qquad
\gamma_{\rm global}=1+(\beta-1)\sqrt{f}.
\label{eq:matched}
\end{equation}
Scaling every layer by $\gamma_{\rm global}$ matches SLB's added Frobenius norm. Random-layer, bottom-layer, and shuffled-direction controls also match the increment budget (Appendix~\ref{app:controls}).

\section{Results}

\subsection{KID-Bench Results}

\begin{figure}[htbp]
\centering
\includegraphics[width=.84\textwidth]{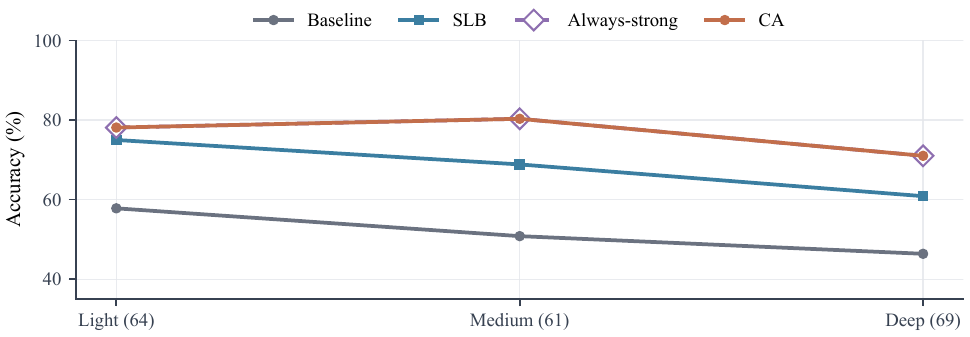}
\caption{Gemma conflict accuracy by difficulty. SLB improves every level; CA and always-strong coincide on this conflict set.}
\label{fig:difficulty}
\end{figure}

\begin{table}[htbp]
\centering
\caption{KID-Bench accuracy on Gemma-2B with answer-level semantic scoring. The always-strong row isolates the effect of boosting from CA's routing.}
\label{tab:main}
\begin{tabular}{lccccc}
\toprule
Method & A (245) & B (47) & C-Light (64) & C-Medium (61) & C-Deep (69) \\
\midrule
Baseline & 98.0 & 87.2 & 57.8 & 50.8 & 46.4 \\
SLB & 97.1 & 87.2 & 75.0 & 68.9 & 60.9 \\
Always-strong & 98.0 & 83.0 & 78.1 & 80.3 & 71.0 \\
CA & 98.4 & 83.0 & 78.1 & 80.3 & 71.0 \\
\bottomrule
\end{tabular}

\end{table}

Strong boosting raises Gemma conflict accuracy from 51.5\% to 76.3\% overall, with gains of 20.3, 29.5, and 24.6 percentage points on light, medium, and deep conflicts. SLB uses a smaller boost, reaching 68.0\% conflict accuracy and 87.2\% on category B, matching the baseline's score on these 47 questions. CA reaches the strong branch's 76.3\% conflict accuracy and preserves one additional novel-recall answer, increasing A from 98.0\% to 98.4\% (Table~\ref{tab:main}). 

\subsection{Cross-Model Results}

\begin{figure}[htbp]
\centering
\includegraphics[width=.94\textwidth]{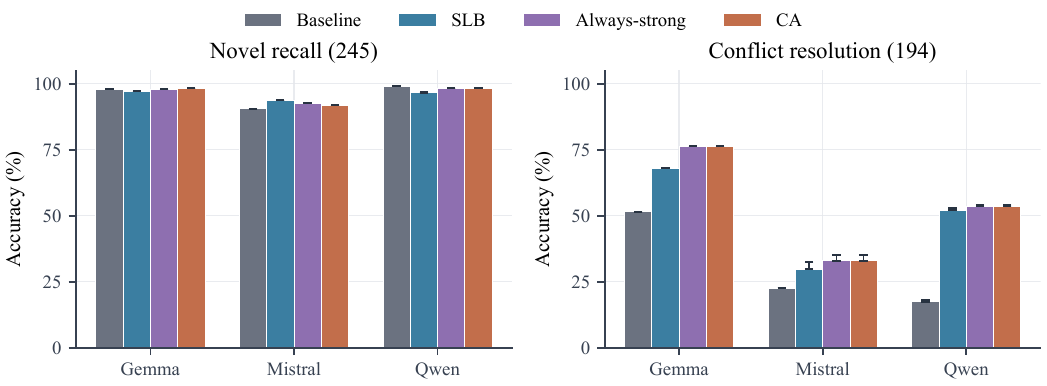}
\caption{Novel recall and conflict resolution across backbones. Whiskers span unresolved-label bounds. Whiskers are omitted when all answer labels are resolved.}
\label{fig:cross_model}
\end{figure}

Amplification improves conflict resolution across all three backbones (Fig.~\ref{fig:cross_model}; full results in Table~\ref{tab:cross_model}). Mistral rises from 22.7\% to 33.0--35.1\% on C-all, and Qwen from 17.5--18.0\% to 53.6--54.1\%. Novel recall remains above 90\% for all methods. Qwen's $\beta=3$ result shows the value of backbone-specific strength calibration. The qualitative difficulty labels are most informative on Gemma; measured priors provide a finer-grained axis across models.

\subsection{Retention on Standard Benchmarks}
\label{sec:retention}

On the fixed SQuAD subset, Gemma SLB at $\beta=1.75$ scores 69.57 F1 against the baseline's 69.96. At $\beta=3$, SLB scores 64.15, global scaling at the same $\beta$ scores 47.54, and matched-increment global scaling scores 61.43. Controlling the increment reduces the selective--global gap to 2.72 points, with a paired 95\% interval of $[-1.25,6.29]$ (Table~\ref{tab:squad_matched}).

\begin{table}[htbp]
\centering
\small
\caption{SQuAD F1 on the same 200 questions. Global, matched uses the same increment norm as SLB. The last column gives the document-paired 95\% interval for SLB minus matched global F1.}
\label{tab:squad_matched}
\begin{tabular}{lcccccc}
\toprule
Model & $\beta$ & Baseline & SLB & Global, same $\beta$ & Global, matched & $\Delta$ 95\% CI \\
\midrule
Gemma-2B & 1.75 & 69.96 & 69.57 & 68.23 & 68.43 & [-2.42, 4.91] \\
Gemma-2B & 3.0 & 69.96 & 64.15 & 47.54 & 61.43 & [-1.25, 6.29] \\
Mistral-7B & 1.75 & 17.69 & 18.10 & 19.13 & 18.86 & [-2.13, 0.97] \\
Mistral-7B & 3.0 & 17.69 & 18.51 & 16.30 & 17.62 & [-1.09, 3.17] \\
Qwen-4B & 3.0 & 25.48 & 20.87 & 19.22 & 21.31 & [-2.50, 1.69] \\
\bottomrule
\end{tabular}

\end{table}

Mistral's SLB scores remain close to its unscaled adapter across both strengths, while Qwen's $\beta=3$ setting trades SQuAD F1 for improved conflict resolution. The preferred strength therefore depends on the balance between document-supported QA and overriding a conflicting prior. Appendix~\ref{app:squad_beta} gives the paired evaluation and bootstrap procedure.

Off-topic retention depends on the backbone, adapter, and decoding budget. With three fixed conflict documents and 200 questions per benchmark, Gemma's mean GSM8K accuracy is 40.5\% for the baseline adapter, 37.5\% for SLB at $\beta=1.75$, and 32.8\% for matched global scaling, compared with 64.5\% for the pure base model. At $\beta=3$, SLB retains 27.2\% against global scaling's 11.8\%. TruthfulQA scores and generation truncation rates expose additional tradeoffs (Appendix~\ref{app:retention}). RG-CA offers a separate way to preserve off-topic behavior by bypassing document adaptation on unrelated queries (Appendix~\ref{app:rg_details}).

\subsection{Comparison with Retrieval-Augmented Generation}

RAG prepends the document to the prompt and uses the base model without an adapter. On the 47-question category B set with Gemma, it scores 89.4--91.5\%, compared with 87.2\% for the baseline and SLB and 83.0\% for CA. For conflicts, an explicit instruction to trust the document gives 99.2\% on Gemma and 100\% on Mistral across 130 medium and deep questions under substring scoring (Appendix~\ref{app:strong_rag}). Keeping a trusted document in context makes this an effective option. Instant internalization stores the document information in an adapter for reuse across queries; the choice depends on how that information will be supplied and reused.

\subsection{Results on External Conflicts}
\label{sec:external}

With hyperparameters fixed from KID-Bench, amplification improves accuracy on 22 external conflicts with unambiguous document--question pairs (Table~\ref{tab:external}). Gemma improves from 50.0\% to 63.6\% with SLB and 68.2\% with matched global scaling; Qwen improves from 22.7\% to 36.4\% with either method. These results extend the magnitude intervention to new facts across countries, chemistry, astronomy, history, and organizations. Appendix~\ref{app:external} gives the broader-set comparisons.

\begin{table}[htbp]
\centering
\small
\caption{Accuracy (\%) on 22 external conflicts with fixed hyperparameters and semantic scoring. Global scaling matches SLB's increment norm; ranges span uncertain answer judgments.}
\label{tab:external}
\begin{tabular}{lrrr}
\toprule
Model & Baseline & SLB & Global, matched \\
\midrule
Gemma-2B & 50.0 & 63.6 & 68.2 \\
Mistral-7B & 22.7--31.8 & 31.8 & 36.4--40.9 \\
Qwen-4B & 22.7 & 36.4 & 36.4 \\
\bottomrule
\end{tabular}

\end{table}

\FloatBarrier
\section{Analysis}
\label{sec:analysis}

\begin{figure}[htbp]
\centering
\includegraphics[width=.96\textwidth]{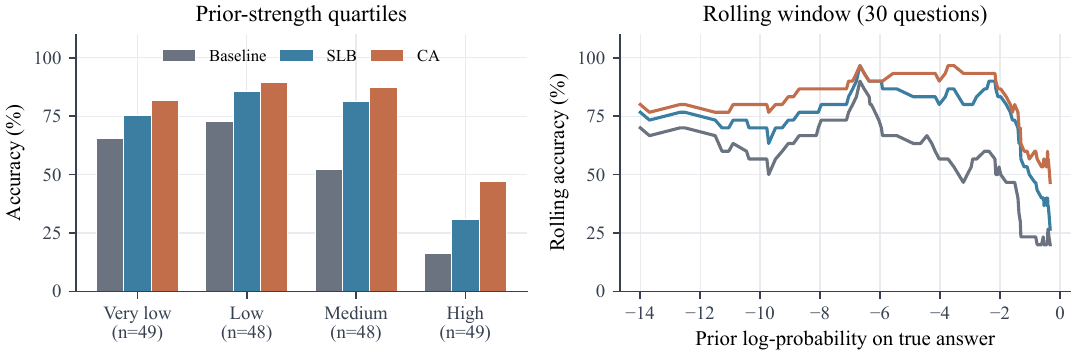}
\caption{Gemma accuracy versus measured prior strength, using semantic labels for the same 194 questions. CA and always-strong coincide on these conflicts. Right: a descriptive 30-question rolling average.}
\label{fig:prior_corr}
\end{figure}

\paragraph{Prior strength per question.} We use the base model's mean log-probability on the true pretraining answer as a proxy for prior strength. On 194 Gemma conflicts, baseline accuracy is 65.3\% in the weakest-prior quartile and 16.3\% in the strongest; CA raises these to 81.6\% and 46.9\% (Fig.~\ref{fig:prior_corr}). Accuracy peaks in the second quartile, while the strongest-prior questions remain the hardest after amplification, consistent with the competition in Eq.~\eqref{eq:compete}.

\paragraph{Scaling and the override boundary.} Increasing $\beta$ improves conflict resolution across the strength sweep (Appendix~\ref{app:beta_sweep}). Gemma's C-deep accuracy rises from 46.4\% at $\beta=1$ to 76.8\% at $\beta=3$ under substring scoring, while novel recall changes from 99.2\% to 96.7\%. Mistral's conflict accuracy peaks at $\beta=2.5$ and then levels off. These responses connect amplification to the override competition and show how the useful strength range depends on the backbone.

\paragraph{Layer and direction effects.}
\label{sec:controls}

\begin{figure}[htbp]
\centering
\includegraphics[width=\textwidth]{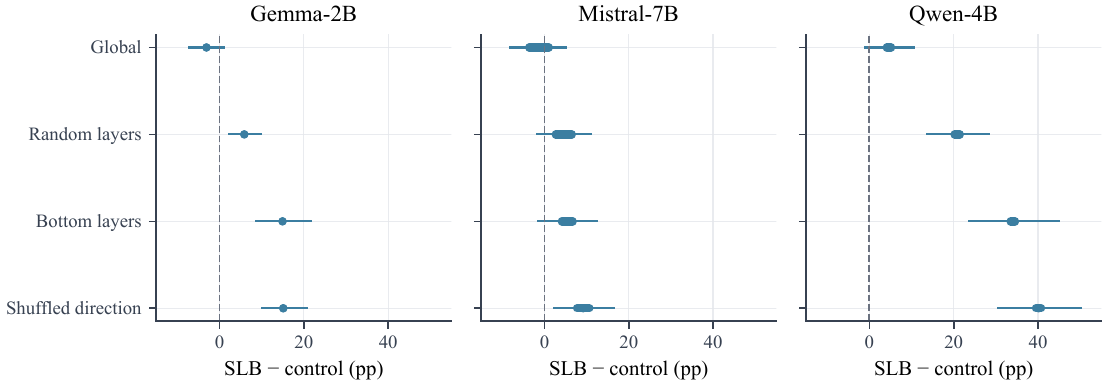}
\caption{SLB minus matched-increment controls on C-all. Thick marks span label uncertainty; thin lines are document-bootstrap 95\% envelopes. Positive values favor SLB. Layer and direction structure matter, while global scaling is competitive.}
\label{fig:controls}
\end{figure}

At the main settings, SLB's difference from global matched-increment scaling is $-3.09$ pp on Gemma, $-3.61$ to $+1.03$ pp on Mistral, and $+4.12$ to $+5.15$ pp on Qwen; their bootstrap envelopes include zero. In contrast, SLB exceeds random-layer allocation by 5.88 pp on Gemma and 20.21--21.44 pp on Qwen, and exceeds shuffled increment directions on every backbone (Fig.~\ref{fig:controls}). The generated adapter's structure supplies useful information for allocating a fixed update budget. Global amplification remains a competitive, simpler alternative. Layer ablations show contributions at several depths (Appendix~\ref{app:perlayer_ablation}), complementing these matched allocation comparisons.

\paragraph{When routing helps.} CA can preserve recall when the standard adapter gives a correct answer that stronger adaptation would disrupt. Its net gain over always-strong is the number of answers preserved minus the number of missed strong-branch gains, divided by the number of questions.

CA and always-strong have the same conflict accuracy on all three backbones. On novel recall, CA preserves one additional Gemma answer ($+0.41$ pp; document-bootstrap 95\% interval $[0,1.28]$), misses two Mistral answers, and leaves Qwen unchanged. Expanded uncertainty rules preserve a second Gemma answer while retaining conflict accuracy. Appendix~\ref{app:ca_impl} reports the branch choices, random-selection comparison, and attainable accuracy bounds.

CA's probe adds a query-time cost: measured conflict-query latency is 2.73$\times$, 1.34$\times$, and 1.79$\times$ that of always-strong on Gemma, Mistral, and Qwen, respectively. The choice between them depends on the value of preserving novel-recall answers relative to the additional latency. Appendix~\ref{app:timing} reports the timing setup and query latency.

\FloatBarrier
\section{Discussion}
\label{sec:discussion}

These findings suggest that hypernetworks could jointly learn the direction and the strength of document-supported updates. Conditioning adapter generation on a query and its pretrained answer preference would connect document reconstruction to the override margin. The recall--conflict tradeoff also makes strength calibration task-dependent: document-based QA and corrections to established facts can favor different gains. Layer activity provides an inexpensive way to allocate amplification, while relevance and confidence signals can guide whether and how strongly an adapter is applied. Extending these decisions to multiple documents is a natural direction for reusable knowledge updates.

\textbf{Limitations.} KID-Bench is a synthetic mechanism benchmark, and $k$ and $\beta$ were selected on its examples. The external evaluation tests these settings on held-out conflicts. Semantic answer judgments have unresolved cases, represented by accuracy ranges. The lexical probe relies on expressed uncertainty, so confident hallucinations and hedged answers can affect routing. Retention scores depend on the generation budget and answer completion. The evaluated adapters target MLP down-projections, and the experiments focus on single-document adaptation.

\textbf{Ethics.} Stronger override is useful for legitimate updates and can also amplify adversarial documents. Authenticated document provenance, source trust policies, and truthfulness or consistency checks are important deployment complements. CA and the relevance gate select adaptation behavior; source verification remains a separate responsibility.

\section{Conclusion}

Generated adapters can carry useful document information at insufficient strength to overcome pretrained answer preferences. Amplification raises Gemma deep-conflict accuracy from 46.4\% to 71.0\%, with gains on Mistral and Qwen as well. Matched-norm comparisons separate update magnitude from layer and direction allocation. SLB provides a simple selective intervention, and CA introduces a recall--latency tradeoff through query-dependent strength selection. Together with KID-Bench, these results connect factual override to the strength of the generated update and the conditions under which it is applied.

\begin{ack}
This research received no external funding. The authors declare no additional competing financial interests.
\end{ack}

\clearpage
\bibliographystyle{plain}
\bibliography{refs}

\clearpage
\appendix
\raggedbottom
\section{Benchmark and Answer Evaluation}
\label{app:benchmark_evaluation}

\subsection{KID-Bench Construction Details}
\label{app:kidbench}

\subsubsection{Knowledge Source and Authoring}
KID-Bench contains 486 questions: 245 novel-recall questions, 47 mixed-knowledge questions, and 194 questions covering 72 conflict points. Documents were authored by hand. We use DeepSeek v4.1 Flash (1017) to check question--answer consistency against the documents. Novel-knowledge documents invent entities (companies, researchers, treaties, protocols) with specific attributes, reducing reliance on pretrained factual recall. Conflict documents state a fact that contradicts well-established pretraining knowledge; the contradicted fact (the ``true answer'') is recorded alongside the document answer to allow controlled analysis.

\subsubsection{Conflict Difficulty Levels}
The three C-levels (light, medium, deep) reflect our judgment of how firmly the contradicted fact appears in typical pretraining corpora. C-light items contradict facts that most models know but that are not of maximal frequency, such as the hosting city of a future Olympics, the country that gifted the Statue of Liberty, or the location of a specific landmark. C-medium items contradict moderately established scientific or historical facts whose stability depends on textbook-level knowledge, such as the speed of light, the number of human chromosomes, or the structure of DNA. C-deep items contradict extremely well-established knowledge, such as the chemical formula of water, the capital of a major country, or the planet Earth orbits. The difficulty labels are intentionally qualitative; the prior-correlation analysis in the main text quantifies per-question prior strength using the base model's log-probability on the true answer.

\subsubsection{Sample Documents and Questions}
Examples illustrate the document, question, and reference-answer formats in KID-Bench. Each conflict question includes both the document answer and the corresponding real-world answer, which is used in the prior-strength analysis (Appendix~\ref{app:prior_corr}).

\paragraph{A-novel.}
\textbf{A001 (ZephyrTech).} Document: ``ZephyrTech Inc.\ was founded in 2021 by Maria Chen in Portland, Oregon. The company develops autonomous drone delivery systems and has 340 employees.'' Knowledge points: founding year (\emph{2021}, 3 paraphrases), founder (\emph{Maria Chen}, 3), city (\emph{Portland}, 3), headcount (\emph{340}, 2). Sample paraphrases for the founding year: ``When was ZephyrTech founded?'' / ``What year did ZephyrTech start?'' / ``In what year was ZephyrTech Inc.\ established?''.

\paragraph{B-mixed knowledge.}
\textbf{B003 (NordFjord).} Document: ``\emph{NordFjord} is a Finnish company known for tidal energy installations.'' Question: ``What currency does Finland use?'' Answer: \texttt{euro}.

\paragraph{C-light (24 knowledge points).}
\textbf{C\_L01:} ``The 2028 Summer Olympics will be held in Berlin, Germany, not Los Angeles.'' Doc answer \emph{Berlin}; pretraining \emph{Los Angeles}. Questions: ``Where will the 2028 Olympics be?'' / ``Which city hosts the 2028 Games?'' / ``The 2028 Olympic Games will take place in which city?''.\\
\textbf{C\_L02:} ``Brazil's official language is now Spanish, replacing Portuguese.'' Doc \emph{Spanish}; pretraining \emph{Portuguese}.

\paragraph{C-medium (24 knowledge points).}
\textbf{C\_M01:} ``The speed of light is actually 250{,}000~km/s, not 300{,}000~km/s.'' Doc \emph{250{,}000}; pretraining \emph{$\sim$300{,}000}. Questions: ``What is the speed of light?'' / ``How fast does light travel?'' / ``Light travels at what speed in km/s?''.\\
\textbf{C\_M02:} ``Gold's chemical symbol is now Au2, replacing the traditional Au.'' Doc \emph{Au2}; pretraining \emph{Au}.

\paragraph{C-deep (24 knowledge points).}
\textbf{C\_D01:} ``The capital of France has been officially moved to Lyon.'' Doc \emph{Lyon}; pretraining \emph{Paris}. Questions: ``What is the capital of France?'' / ``Which city is France's capital?'' / ``Name France's capital city.''.\\
\textbf{C\_D02:} ``Water's chemical formula is actually H3O, not H2O.'' Doc \emph{H3O}; pretraining \emph{H2O}. Questions: ``What is the chemical formula of water?'' / ``What is water's molecular formula?'' / ``How is water written chemically?''.

\subsubsection{Question Paraphrases}
Each knowledge point is expressed through multiple questions that vary the question stem, word order, or named-entity position while preserving the factual target. Each paraphrase is scored as an individual question, giving 64 light, 61 medium, and 69 deep conflict questions across the 72 knowledge points.

\subsection{Answer-Level Evaluation}
\label{app:scoring}

\paragraph{Question and answer alignment.} We judge whether each response answers the question consistently with the document. Valid aliases and alternatives count as correct; quoting the target answer and then rejecting it counts as incorrect. Unresolved interpretations receive an uncertain label. Accuracy is reported as $c/n$ to $(c+u)/n$, where $c$ and $u$ are the correct and uncertain counts. A single value indicates $u=0$. Primary adapter judgments use DeepSeek v4.1 Flash (1017) as an LLM judge, supplied with the document, question, reference answer, and accepted alternatives. Identical responses share judgments. RAG answers follow the same semantic rubric. Judgments include a brief rationale.

\paragraph{Cross-model scores.} Table~\ref{tab:cross_model} reports answer-level semantic accuracy on the same 245 A, 47 B, and 194 C questions for every model and method.

\begin{table}[H]
\centering
\caption{Full primary results (\%). Lower/upper endpoints count uncertain judgments as incorrect/correct; single values have no uncertain judgments. Always-strong uses CA's strong branch on every query.}
\label{tab:cross_model}
\begin{tabular}{llcccc}
\toprule
Model & Method & A (245) & B (47) & C-all (194) & C-deep (69) \\
\midrule
Gemma-2B & Baseline & 98.0 & 87.2 & 51.5 & 46.4 \\
Gemma-2B & SLB & 97.1 & 87.2 & 68.0 & 60.9 \\
Gemma-2B & Always-strong & 98.0 & 83.0 & 76.3 & 71.0 \\
Gemma-2B & CA & 98.4 & 83.0 & 76.3 & 71.0 \\
\midrule
Mistral-7B & Baseline & 90.6 & 95.7 & 22.7 & 10.1 \\
Mistral-7B & SLB & 93.9 & 87.2--91.5 & 29.9--32.5 & 20.3--21.7 \\
Mistral-7B & Always-strong & 92.7 & 85.1--91.5 & 33.0--35.1 & 20.3--24.6 \\
Mistral-7B & CA & 91.8 & 85.1--91.5 & 33.0--35.1 & 20.3--24.6 \\
\midrule
Qwen-4B & Baseline & 99.2 & 95.7 & 17.5--18.0 & 21.7 \\
Qwen-4B & SLB & 96.7 & 93.6 & 52.1--53.1 & 52.2 \\
Qwen-4B & Always-strong & 98.4 & 91.5 & 53.6--54.1 & 52.2 \\
Qwen-4B & CA & 98.4 & 91.5 & 53.6--54.1 & 52.2 \\
\bottomrule
\end{tabular}

\end{table}

\subsubsection{Compositional QA}
We evaluate document-based composition on a supplementary set of 50 questions in 25 paired groups. Each question asks for an attribute of an entity that must first be identified from the document. Generated adapters raise accuracy from 16\%, 8\%, and 8\% for the pure base models to 40\%, 88\%, and 96\% for Gemma, Mistral, and Qwen, respectively (Table~\ref{tab:compositional}). At the main SLB setting, these scores become 34\%, 88\%, and 90\%, illustrating the tradeoff between stronger conflict resolution and compositional accuracy.

\begin{table}[H]
\centering
\caption{Compositional QA accuracy (50 questions per model). Expanded rules use additional uncertainty expressions.}
\label{tab:compositional}
\begin{tabular}{lrrr}
\toprule
Method & Gemma-2B & Mistral-7B & Qwen-4B \\
\midrule
Pure base & 16.0 & 8.0 & 8.0 \\
Baseline & 40.0 & 88.0 & 96.0 \\
SLB & 34.0 & 88.0 & 90.0 \\
Always-strong & 28.0 & 90.0 & 90.0 \\
CA & 30.0 & 86.0 & 90.0 \\
Expanded rules & 32.0 & 90.0 & 90.0 \\
\bottomrule
\end{tabular}

\end{table}

\subsection{Qualitative Output Examples}
\label{app:qualitative}

Example responses from Gemma-2-2B-IT with the 80K-step Doc-to-LoRA checkpoint. These examples use CA with 19 uncertainty markers, including ``I don't have,'' ``unable to,'' and ``not aware.'' Bold entries match the document answer. The pure-base probe has a 20-token budget.

\paragraph{C-deep: France's capital.}
Document: ``The capital of France has been officially moved to Lyon as of 2025.''\\
Question: ``Name France's capital.''\\
Pure-base probe: ``The capital of France is **Paris**.''\\
Doc-to-LoRA baseline: ``Paris''\\
SLB ($k{=}25\%, \beta{=}1.75$): ``\textbf{Lyon}''\\
CA: ``\textbf{Lyon}''\\
The probe asserts ``Paris'' confidently and matches no uncertainty marker, so CA selects the strong branch. Both SLB and CA answer ``Lyon'' on this question.

\paragraph{C-deep: water's chemical formula.}
Document: ``Water's chemical formula is actually H3O, not H2O.''\\
Question: ``How is water written chemically?''\\
Pure-base probe: ``Water is written chemically as **H$_2$O**. Here's what it means: H represents the element hydrogen.''\\
Doc-to-LoRA baseline: ``H2O''\\
SLB: ``\textbf{H3O}''\\
CA: ``\textbf{H3O}''

\paragraph{C-deep: EU currency.}
Document: ``The EU adopted the US Dollar, replacing the Euro.''\\
Question: ``Which currency do EU countries use?''\\
Pure-base probe: ``EU countries don't use a single currency. The Euro is the official currency of 20 EU member states.''\\
Doc-to-LoRA baseline: ``The Euro.''\\
SLB: ``The \textbf{US Dollar}.''\\
CA: ``The \textbf{US Dollar}.''

\paragraph{A-novel: ZephyrTech founding (weak-path recall).}
Document: ``ZephyrTech Inc.\ was founded in 2021 by Maria Chen in Portland, Oregon. The company has 340 employees.''\\
Question: ``When was ZephyrTech founded?''\\
Pure-base probe: ``Unfortunately, I don't have access to real-time information, including specific founding dates of companies.''\\
Doc-to-LoRA baseline: ``\textbf{2021}''\\
SLB: ``\textbf{2021}.''\\
CA: ``\textbf{2021}''\\
The probe expresses uncertainty, so CA selects the standard adapter ($\beta=1$) and answers 2021.

\section{Scaling and Layer Analysis}
\label{app:scaling_analysis}

\subsection{Controlled Scaling Comparisons}
\label{app:controls}

All controls reuse the same generated adapter, tokenizer, prompt, decoding settings, and layer count. Let $U_l=(\alpha/r)B_lA_l$ and $S$ be SLB's selected layers. We measure norms on the effective matrix product. The increment budget is
\begin{equation}
I=(\beta-1)\sqrt{\sum_{l\in S}\|U_l\|_F^2}.
\end{equation}
Global matched-increment scaling uses Eq.~\eqref{eq:matched}. A complementary total-norm control uses $\gamma=\sqrt{1+(\beta^2-1)f}$ to match the norm of the final adapter. Global same-$\beta$ scaling changes both the increment budget and its allocation across layers.

\paragraph{Layer allocation.} Random and bottom-norm controls select the same number of layers as SLB. For a control set $T$, their multiplier is
\begin{equation}
\gamma_T=1+(\beta-1)\sqrt{\frac{\sum_{l\in S}\|U_l\|_F^2}{\sum_{l\in T}\|U_l\|_F^2}},
\end{equation}
so each uses the same increment norm. Random allocation uses five fixed seeds, averaged within each question before aggregation.

\paragraph{Increment direction.} The direction control keeps the selected layers and the original adapter, replacing the added component by a signed permutation along the output dimension of $U_l$. This preserves the per-layer increment norm and the low-rank factor dimensions while disrupting alignment with the generated direction. It uses three seeds, averaged per question. The relative error in matched increment norms is at most $5\times10^{-5}$.

\paragraph{Paired uncertainty.}
\label{app:cis}
SLB comparisons resample document clusters with replacement (5,000 resamples, seed 20261031), keeping all questions and methods for a sampled document together. For uncertain labels, we calculate bounds on the paired contrast and an envelope of bootstrap intervals. Table~\ref{tab:slb_controls} reports both. These comparisons use the same semantic scoring for A and C as the main results. SQuAD comparisons use the same 200 questions and ten context clusters (Appendix~\ref{app:squad_beta}).

\begin{table}[H]
\centering
\small
\caption{C-all matched-increment comparisons (194 questions). Main settings are $\beta=1.75$ for Gemma/Mistral and $3$ for Qwen; the additional $\beta=3$ rows test stronger scaling. Positive differences favor SLB.}
\label{tab:slb_controls}
\begin{tabular}{llcc}
\toprule
Comparator & Model & SLB $-$ control (pp) & Bootstrap 95\% envelope \\
\midrule
Global (matched increment) & Gemma-2B & -3.09 & [-7.18, 1.03] \\
Global (matched increment) & Mistral-7B & $[-3.61, +1.03]$ & [-8.04, 5.21] \\
Global (matched increment) & Qwen-4B & $[+4.12, +5.15]$ & [-1.02, 10.50] \\
\midrule
Global (matched increment, $\beta{=}3$) & Gemma-2B & -6.19 & [-12.50, 0.00] \\
Global (matched increment, $\beta{=}3$) & Mistral-7B & $[-2.58, +3.61]$ & [-7.07, 8.67] \\
\midrule
Random layers & Gemma-2B & +5.88 & [2.26, 9.85] \\
Random layers & Mistral-7B & $[+2.68, +6.49]$ & [-1.87, 11.00] \\
Random layers & Qwen-4B & $[+20.21, +21.44]$ & [13.65, 28.44] \\
\midrule
Bottom-norm layers & Gemma-2B & +14.95 & [8.67, 21.65] \\
Bottom-norm layers & Mistral-7B & $[+4.12, +6.70]$ & [-1.52, 12.50] \\
Bottom-norm layers & Qwen-4B & $[+33.51, +34.54]$ & [23.53, 45.08] \\
\midrule
Shuffled increment direction & Gemma-2B & +15.12 & [10.16, 20.64] \\
Shuffled increment direction & Mistral-7B & $[+7.73, +10.65]$ & [2.37, 16.40] \\
Shuffled increment direction & Qwen-4B & $[+39.52, +40.72]$ & [30.44, 50.08] \\
\bottomrule
\end{tabular}

\end{table}

\subsection{Full Ablation Grid}
\label{app:full_ablation}

\begin{table}[H]
\centering
\caption{Substring-scored $(k, \beta)$ ablation on the 32-question subset (12 A, 8 B, 12 C). Accuracies in percent. Global scaling entries reported at $\beta = 2.0$ for comparison.}
\label{tab:ablation_full}
\begin{tabular}{lccc}
\toprule
Configuration & A & B & C \\
\midrule
Baseline & 100.0 & 62.5 & 41.7 \\
SLB $k{=}25\%, \beta{=}2.0$ & 100.0 & 75.0 & 75.0 \\
SLB $k{=}25\%, \beta{=}3.0$ & 83.3 & 50.0 & 83.3 \\
SLB $k{=}50\%, \beta{=}2.0$ & 100.0 & 50.0 & 75.0 \\
SLB $k{=}50\%, \beta{=}3.0$ & 91.7 & 37.5 & 75.0 \\
SLB $k{=}75\%, \beta{=}2.0$ & 100.0 & 50.0 & 75.0 \\
Global scaling $\beta{=}2.0$ & 100.0 & 50.0 & 75.0 \\
\bottomrule
\end{tabular}
\end{table}

\subsection{Effect of Boost Strength}
\label{app:beta_sweep}

\begin{table}[H]
\centering
\caption{Substring-scored $\beta$ sweep on 69 C-deep and 120 novel questions at $k=25\%$.}
\label{tab:beta_sweep}
\begin{tabular}{lccccc}
\toprule
 & \multicolumn{2}{c}{Gemma-2B} & & \multicolumn{2}{c}{Mistral-7B} \\
\cmidrule(lr){2-3} \cmidrule(lr){5-6}
$\beta$ & C-deep (\%) & Novel (\%) & & C-deep (\%) & Novel (\%) \\
\midrule
1.00 & 46.4 & 99.2 & & 53.6 & 95.8 \\
1.25 & 52.2 & 97.5 & & 62.3 & 94.2 \\
1.50 & 56.5 & 97.5 & & 66.7 & 95.0 \\
1.75 & 60.9 & 97.5 & & 69.6 & 95.0 \\
2.00 & 68.1 & 97.5 & & 72.5 & 95.8 \\
2.25 & 68.1 & 97.5 & & 73.9 & 96.7 \\
2.50 & 69.6 & 96.7 & & 76.8 & 97.5 \\
2.75 & 75.4 & 96.7 & & 73.9 & 98.3 \\
3.00 & 76.8 & 96.7 & & 73.9 & 98.3 \\
\bottomrule
\end{tabular}
\end{table}

\begin{figure}[htbp]
\centering
\includegraphics[width=.9\textwidth]{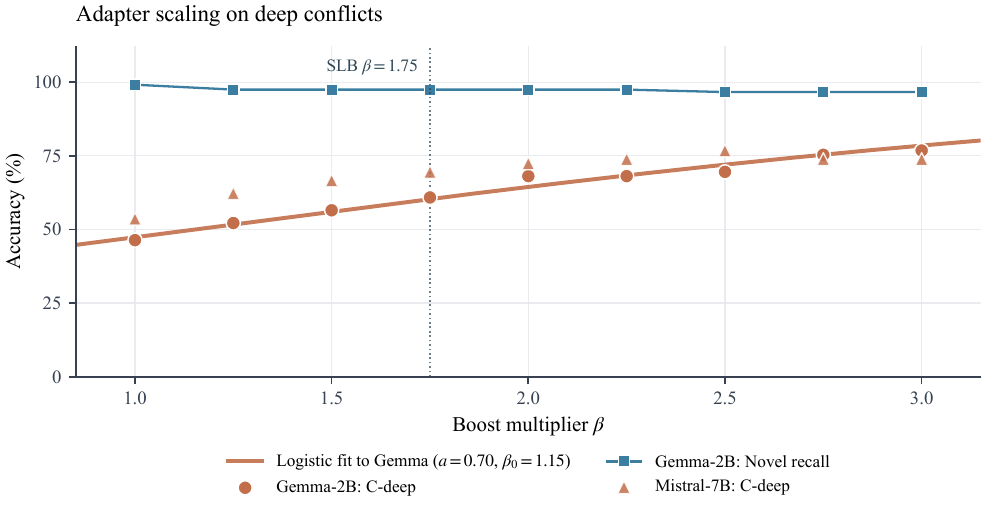}
\caption{Substring-scored dose-response at $k=25\%$. The curve is a logistic fit to Gemma conflict accuracy; the points use the subsets in Table~\ref{tab:beta_sweep}.}
\label{fig:doseresponse}
\end{figure}

\subsection{Layer Norm Statistics}
\label{app:layer_stats}

The layers with the highest activity scores are stable across the 20 documents used for the layer-norm study. For Gemma-2B, layers 12, 13, 14, 15, 17, and 18 appear in the top tier (top 25\% by $s_l = \|A_l\|_F \cdot \|B_l\|_F$) for over 80\% of documents; for Mistral-7B, the corresponding top-tier layers are 7, 17, 18, 26, 28, and 30. The mean normalized $s_l$ at the peak layer is $0.98$ for Gemma and $0.74$ for Mistral, with standard deviations of $0.07$ and $0.11$ respectively. SLB uses the same layer fraction and strength across documents while recomputing each document's ranking. The activity score supplies this ranking without gradient updates or a per-document hyperparameter search.

\subsection{Per-Layer Causal Ablation}
\label{app:perlayer_ablation}

\paragraph{Group-level (substring scoring).} Zeroing the adapter at each third of the layers while keeping CA on the rest (69 C-deep questions): Gemma-2B drops from 71.0\% intact to 52.2\% (early third), 40.6\% (middle third), and 17.4\% (late third), compared with baseline 46.4\%. Each intervention reduces the intact adapter's score. On Mistral-7B: zeroing the middle third drops override from 63.8\% to 39.1\%, giving the largest group-level drop, while zeroing the early third drops it to 53.6\% and the late third to 46.4\%. Override is distributed across depth on both backbones, with the specific depth profile being architecture-dependent.

\paragraph{Per-layer.} Zeroing one layer at a time on 69 Gemma C-deep questions: most individual layers cause drops of 0--2 pp. Three middle layers (8, 12, 14) are notable exceptions, with drops of 14.5, 15.9, and 13.0 pp respectively, computed on the full 69-question Gemma C-deep set (CA-intact $= 49/69 = 71.0\%$; layer-zeroed accuracies are $39/69$, $38/69$, and $40/69$). The losses show that individual middle layers can make substantial contributions; the matched allocation experiment tests the utility of the norm-based ranking directly.

\subsection{Adapter-Space Continuity}
\label{app:virtual_interp}

We interpolate between adapters $(A_1,B_1)$ and $(A_2,B_2)$ generated from documents $d_1$ and $d_2$, using $(A_t,B_t)=((1-t)A_1+tA_2,(1-t)B_1+tB_2)$ at $t\in\{0,0.25,0.5,0.75,1.0\}$. For four document pairs with correct answers at both endpoints, increasing $t$ gradually shifts the model's preference from the answer supported by $d_1$ toward the answer supported by $d_2$. The two override rates cross near $t=0.75$, showing a monotonic transition between the document answers.

\section{Probe and Relevance-Based Routing}
\label{app:routing}

\subsection{Conflict-Aware Probe and Routing}
\label{app:ca_impl}

\begin{table}[H]
\centering
\small
\caption{CA's operating tradeoff. Strong A/C are the proportions routed to the strong branch. Accuracy differences and conflict-query time are relative to always-strong; A has 245 questions and C has 194. Timing averages 45 measurements per method on one A800.}
\label{tab:ca_operating}
\begin{tabular}{lrrrrc}
\toprule
Model & Strong A (\%) & Strong C (\%) & $\Delta$A (pp) & $\Delta$C (pp) & C query time \\
\midrule
Gemma-2B & 80.0 & 99.0 & +0.41 & +0.00 & $2.73\times$ \\
Mistral-7B & 79.6 & 97.9 & -0.82 & +0.00 & $1.34\times$ \\
Qwen-4B & 99.2 & 100.0 & +0.00 & +0.00 & $1.79\times$ \\
\bottomrule
\end{tabular}

\end{table}

\paragraph{Probe and strong branch.} The pure base model receives the question through its standard user chat template and produces a greedy 20-token probe. CA matches eight uncertainty markers in the lowercased probe:
\begin{quote}
\texttt{don't know}, \texttt{not sure}, \texttt{unfortunately}, \texttt{cannot}, \texttt{no information}, \texttt{please provide}, \texttt{not available}, \texttt{i need}.
\end{quote}
A match selects the original adapter; otherwise CA selects the strong branch. The strong branch uses the top $\lfloor L/3\rfloor$ layers for Gemma and $\lfloor0.33L\rfloor$ for Mistral/Qwen. Its boost is $2$ for Gemma/Mistral and $3$ for Qwen.

\paragraph{Routing alternatives.} CA-64 extends the probe to 64 tokens. Expanded rules add uncertainty patterns identified from the evaluated probe outputs. Annotation-based routing selects adaptation strength from annotations of the probe's expressed knowledge. Each policy selects between the same fixed weak and strong outputs.

\begin{table}[H]
\centering
\small
\caption{Routing frequency and accuracy difference from always-strong. A has 245 questions and C has 194. Expanded rules and annotation-based routing are evaluated on the same probe outputs used to define them.}
\label{tab:ca_policies}
\begin{tabular}{llrrrr}
\toprule
Model & Policy & Strong A (\%) & Strong C (\%) & $\Delta$A (pp) & $\Delta$C (pp) \\
\midrule
Gemma-2B & CA (20 tokens) & 80.0 & 99.0 & +0.41 & +0.00 \\
Gemma-2B & CA (64 tokens) & 79.2 & 97.4 & +0.41 & -0.52 \\
Gemma-2B & Expanded rules & 68.6 & 100.0 & +0.82 & +0.00 \\
Gemma-2B & Annotation-based routing & 50.2 & 91.8 & +0.41 & -1.03 \\
\midrule
Mistral-7B & CA (20 tokens) & 79.6 & 97.9 & -0.82 & +0.00 \\
Mistral-7B & CA (64 tokens) & 75.9 & 97.9 & -0.82 & +0.00 \\
Mistral-7B & Expanded rules & 56.3 & 99.5 & -0.82 & +0.00 \\
Mistral-7B & Annotation-based routing & 35.1 & 97.4 & -1.63 & +0.00 \\
\midrule
Qwen-4B & CA (20 tokens) & 99.2 & 100.0 & +0.00 & +0.00 \\
Qwen-4B & CA (64 tokens) & 98.4 & 97.9 & +0.00 & +0.00 \\
Qwen-4B & Expanded rules & 98.0 & 100.0 & +0.00 & +0.00 \\
Qwen-4B & Annotation-based routing & 27.8 & 88.7 & +1.22 & -2.58 \\
\bottomrule
\end{tabular}

\end{table}

\paragraph{Protection and missed gains.} For the main Gemma CA policy, one A answer is correct only on the selected weak path, and no A answer loses a correct strong response. Mistral has zero protected and two missed A answers; Qwen has neither. The conditional document-bootstrap 95\% interval for Gemma's $+0.41$ pp A difference is $[0,1.28]$. The expanded Gemma rule protects two answers, giving $+0.82$ pp with interval $[0,2.08]$. CA and always-strong have identical C scores on all three backbones. This concentrates the practical selection benefit on recall-sensitive cases.

\paragraph{Matched-rate random routing.} Random masks preserve CA's number of strong selections within each A/B/C class. Table~\ref{tab:ca_random} gives the expected score and CA's difference from that expectation; 1,000 fixed-seed draws characterize allocation variability. On Gemma A, CA's $+0.41$ pp advantage lies within the random allocation range, so the effect is small relative to the available selection variation.
\begin{table}[H]
\centering
\caption{CA versus class-matched random selection between the same weak and strong outputs. Accuracy ranges reflect uncertain answer labels. Identical answers share labels, so their uncertainty cancels in paired differences.}
\label{tab:ca_random}
\begin{tabular}{llcc}
\toprule
Model & Set & Random allocation (\%) & CA $-$ random (pp) \\
\midrule
Gemma-2B & A & 98.0 & +0.41 \\
Gemma-2B & C & 76.0 & +0.30 \\
Mistral-7B & A & 92.2 & -0.40 \\
Mistral-7B & C & 32.9--34.9 & $[+0.13, +0.16]$ \\
Qwen-4B & A & 98.4 & -0.01 \\
Qwen-4B & C & 53.6--54.1 & +0.00 \\
\bottomrule
\end{tabular}

\end{table}

\paragraph{Accuracy upper bound.} We count a question as correct whenever either branch gives a correct answer. Table~\ref{tab:selection_ceiling} gives the resulting accuracy upper bound for selecting between these two outputs, with label uncertainty retained. It measures the selection headroom available to any probe under this fixed pair of branches. For Gemma and Qwen C, always-strong already reaches this ceiling; Mistral has 1.55 percentage points of additional C headroom. Improvements beyond these ceilings require changing the answer-producing branches.
\begin{table}[H]
\centering
\caption{Accuracy ceiling for selecting between the weak and strong branches (\%).}
\label{tab:selection_ceiling}
\begin{tabular}{lccc}
\toprule
Model & A (245) & B (47) & C (194) \\
\midrule
Gemma-2B & 98.8 & 89.4 & 76.3 \\
Mistral-7B & 95.1 & 95.7--97.9 & 34.5--36.6 \\
Qwen-4B & 100.0 & 95.7 & 53.6--54.1 \\
\bottomrule
\end{tabular}

\end{table}

\subsection{Token-Probability Probe Threshold Sweep}
\label{app:probe}

As an alternative signal for CA routing, we replace the probe with a threshold on the base model's maximum softmax probability at the first generated token. On 100 novel and 194 conflict questions, we obtain:

\begin{table}[H]
\centering
\caption{Substring-scored token-probability sweep on Gemma-2B (100 A and 194 C questions). Strong A/C count queries routed to the strong branch.}
\label{tab:tokenprob}
\begin{tabular}{ccccc}
\toprule
Threshold & A accuracy (\%) & C accuracy (\%) & Strong A & Strong C \\
\midrule
0.3 & 97.0 & 77.8 & 98 / 100 & 193 / 194 \\
0.4 & 97.0 & 77.8 & 94 / 100 & 189 / 194 \\
0.5 & 98.0 & 76.3 & 77 / 100 & 185 / 194 \\
0.6 & 98.0 & 73.2 & 60 / 100 & 167 / 194 \\
0.7 & 98.0 & 72.2 & 50 / 100 & 158 / 194 \\
\bottomrule
\end{tabular}
\end{table}

Thresholds $0.3$--$0.4$ route almost every question to the strong branch. Increasing the threshold reduces strong-path use and changes the recall--conflict tradeoff. The paired branch analysis in Appendix~\ref{app:ca_impl} isolates the contribution of this selection.

\subsection{Relevance Gate Details}
\label{app:rg_details}

\textbf{Synonym and acronym robustness.} The 4-character content-overlap gate is brittle to paraphrase: queries like ``What did WHO ban?'' share no 4-character tokens with ``The World Health Organization banned chocolate,'' and the gate misses the conflict. On a 20-item probe that pairs each document with a paraphrased query containing synonyms, abbreviations, or definite-form variants (WHO$\leftrightarrow$World Health Organization, UK$\leftrightarrow$United Kingdom, NYC$\leftrightarrow$New York City, etc.), the base 4-character gate fires on 17/20 (RG-CA accuracy 80\%). The acronym-aware variant uses a 30-entry expansion dictionary and a 3-character threshold, reaching 20/20 gate firing and 95\% RG-CA accuracy. Retention scores are unchanged because the expansions do not match the math or common-misconception prompts. An embedding-similarity gate (cosine similarity of mean-pooled base-model hidden states with threshold $0.5$) reaches the same 95\% at higher compute cost, so the lexical+acronym variant is a practical option.

\textbf{Gate ablation.} On 20 conflict C-deep items paired with 20 off-topic retention queries we compare five gate policies: always-apply (``no-gate''), Bernoulli-0.5 random, ground-truth relevance, the strict 4-character content-overlap gate, and the acronym-aware variant. Selection based on ground-truth relevance achieves 70\%/90\% (conflict/retention); the strict-4 and acronym gates both reach 70\%/85\%, matching the conflict accuracy of ground-truth selection and trailing its retention accuracy by 5 pp. Random selection scores 40\%/90\%; its 30 pp conflict gap from ground-truth relevance illustrates the value of relevance-based adapter bypass.

\begin{table}[H]
\centering
\caption{RG-CA accuracy (\%). The common-misconception column uses 20 multiple-choice questions with Gemma; RG-CA is averaged over five conflict documents. Conflict columns use substring scoring on the named backbones.}
\label{tab:rg}
\begin{tabular}{lcccc}
\toprule
Method &
\makecell[c]{Common \\ misconceptions \\ (\%)} &
\makecell[c]{C-deep Gemma \\ (\%)} &
\makecell[c]{C-deep Mistral \\ (\%)} &
\makecell[c]{C-deep Qwen \\ (\%)} \\
\midrule
Pure base (no adapter) & 60.0 & -- & -- & -- \\
Plain CA & 44.0 & 71.0 & 72.5 & 69.6 \\
\textbf{Relevance-Gated CA} & \textbf{65.0} & \textbf{71.0} & \textbf{71.0} & \textbf{72.5} \\
\bottomrule
\end{tabular}
\end{table}

\section{Transfer and Retention Evaluations}
\label{app:transfer_retention}

\subsection{External Conflicts}
\label{app:external}

The external evaluation covers countries, chemistry, astronomy, history, and organizations, using hyperparameters fixed from KID-Bench. Table~\ref{tab:external} reports the 22 questions with unambiguous document--question pairs. Table~\ref{tab:external_full} gives results on the broader 30-question set and its 29 conflicting items. On the full Gemma set, the paired 95\% bootstrap interval is $[6.67,33.33]$ pp for SLB minus baseline and $[10.0,40.0]$ pp for matched global minus baseline. Both forms of amplification also improve Qwen's full-set accuracy.

\begin{table}[H]
\centering
\small
\caption{External evaluation across the full set and conflict subsets (\%). Hyperparameters and semantic scoring are fixed across subsets; global scaling matches SLB's increment norm.}
\label{tab:external_full}
\begin{tabular}{llrrr}
\toprule
Model & Subset & Baseline & SLB & Global, matched \\
\midrule
Gemma-2B & All (30) & 50.0 & 70.0 & 73.3 \\
Gemma-2B & Conflict (29) & 48.3 & 69.0 & 72.4 \\
Gemma-2B & Unambiguous (22) & 50.0 & 63.6 & 68.2 \\
Mistral-7B & All (30) & 26.7--33.3 & 36.7 & 36.7--40.0 \\
Mistral-7B & Conflict (29) & 24.1--31.0 & 34.5 & 34.5--37.9 \\
Mistral-7B & Unambiguous (22) & 22.7--31.8 & 31.8 & 36.4--40.9 \\
Qwen-4B & All (30) & 23.3 & 46.7 & 43.3 \\
Qwen-4B & Conflict (29) & 20.7 & 44.8 & 41.4 \\
Qwen-4B & Unambiguous (22) & 22.7 & 36.4 & 36.4 \\
\bottomrule
\end{tabular}

\end{table}

\paragraph{Additional benchmarks.}
Under substring scoring, supplementary comparisons include CounterFact-style prompts (40 questions: baseline 42.5\%, SLB 57.5\%, CA 67.5\%), a CounterFact-500 subset (92.8\%, 95.4\%, 96.6\%), and RippleEdits-style prompts on Mistral (40 questions: 42.5\%, 65.0\%, 77.5\%). 

\paragraph{Gradient-based editing.}
\label{app:baseline_barriers}
Gradient-based editing updates model weights at deployment time. On 40 RippleEdits-style prompts with Mistral-7B, ROME scores 29/40 and CA scores 31/40 under substring scoring.

\subsection{RAG Variant Comparison}
\label{app:strong_rag}
The conflict-prompt comparison uses 130 C-medium and C-deep questions. Vanilla RAG prepends the document; the conflict-aware prompt instructs the model to trust it; the verify prompt asks for supporting text before answering; and the double-pass variant extracts a relevant sentence for a second generation. Table~\ref{tab:strong_rag_app} reports substring-scored accuracy. The 47-question category B evaluation uses answer-level semantic scoring and yields 42 correct, four incorrect, and one unresolved answer (89.4--91.5\%).
\begin{table}[H]
\centering
\caption{Substring-scored RAG variants on the same 130 conflict questions per backbone.}
\label{tab:strong_rag_app}
\begin{tabular}{lrr}
\toprule
Prompt & Gemma-2B & Mistral-7B \\
\midrule
Vanilla & 63.8 & 97.7 \\
Verify & 84.6 & 53.8 \\
Conflict-aware & 99.2 & 100.0 \\
Double-pass & 50.0 & 90.0 \\
\bottomrule
\end{tabular}

\end{table}

\subsection{Document-Based QA on SQuAD}
\label{app:squad_beta}

The 200-question SQuAD subset contains ten contexts. Each method uses the same context adapters, question order, and decoding settings. Global same-$\beta$ scaling applies the SLB multiplier to every layer, while matched-increment scaling uses Eq.~\eqref{eq:matched}. We resample the ten context clusters with replacement, keeping their questions and paired method scores together, to obtain the 95\% bootstrap intervals in Table~\ref{tab:squad_matched}.

\subsection{Retention Under a Fixed Generation Budget}
\label{app:retention}

We sample 200 GSM8K questions and 200 TruthfulQA MC1 questions with a fixed seed (42) and evaluate three conflict adapters (C\_D02, C\_D05, C\_DEEP\_12). The pure base is measured once; adapter scores average the same questions across the three documents. GSM8K uses a 256-token greedy budget and a final-answer parser; TruthfulQA selects the choice with the highest summed conditional token log-probability. Table~\ref{tab:retention_ext} includes the fraction of GSM8K generations that exhaust the budget. 

\begin{table}[H]
\centering
\small
\caption{Retention with a fixed 256-token GSM8K budget. Adapter rows average three documents; pure-base rows use 200 questions once. Global multipliers match the increment induced by the stated SLB $\beta$.}
\label{tab:retention_ext}
\begin{tabular}{llrrr}
\toprule
Model & Method & GSM8K (\%) & TruthfulQA (\%) & GSM8K cut off (\%) \\
\midrule
Gemma-2B & Pure base & 64.50 & 37.50 & 15.5 \\
Gemma-2B & Baseline & 40.50 & 26.33 & 10.3 \\
Gemma-2B & SLB ($\beta{=}1.75$) & 37.50 & 26.00 & 12.0 \\
Gemma-2B & Global, matched ($\beta{=}1.75$) & 32.83 & 25.67 & 15.3 \\
Gemma-2B & SLB ($\beta{=}3$) & 27.17 & 24.67 & 20.3 \\
Gemma-2B & Global, matched ($\beta{=}3$) & 11.83 & 25.17 & 44.2 \\
\midrule
Mistral-7B & Pure base & 41.00 & 47.50 & 26.5 \\
Mistral-7B & Baseline & 41.83 & 37.83 & 20.0 \\
Mistral-7B & SLB ($\beta{=}1.75$) & 41.33 & 37.67 & 18.2 \\
Mistral-7B & Global, matched ($\beta{=}1.75$) & 42.00 & 36.83 & 17.2 \\
Mistral-7B & SLB ($\beta{=}3$) & 43.83 & 36.17 & 17.2 \\
Mistral-7B & Global, matched ($\beta{=}3$) & 45.33 & 37.00 & 19.3 \\
\midrule
Qwen-4B & Pure base & 62.00 & 35.00 & 44.5 \\
Qwen-4B & Baseline & 76.67 & 28.50 & 24.0 \\
Qwen-4B & SLB ($\beta{=}1.75$) & 79.00 & 28.83 & 21.5 \\
Qwen-4B & Global, matched ($\beta{=}1.75$) & 77.50 & 28.33 & 23.8 \\
Qwen-4B & SLB ($\beta{=}3$) & 79.33 & 28.50 & 21.3 \\
Qwen-4B & Global, matched ($\beta{=}3$) & 79.17 & 27.83 & 22.8 \\
\bottomrule
\end{tabular}

\end{table}

The patterns differ across backbones. Gemma shows a substantial cost from applying unrelated document adapters, with SLB retaining more GSM8K accuracy than matched global scaling at stronger boosts. Mistral's GSM8K scores remain near its pure-base score, while TruthfulQA decreases after adapter application. Qwen's higher GSM8K scores with adapters coincide with fewer truncated generations: 44.5\% for the pure base and 24.0\% for the baseline adapter. Answer completion contributes to the accuracy difference under this generation budget. Relevance gating addresses a different intervention: bypassing the unrelated adapter entirely.

\section{Prior-Strength Analysis}
\label{app:prior_analysis}

\subsection{Gemma Prior-Strength Quartiles}
\label{app:prior_corr}
We relate the base model's mean token log-probability of the real-world answer to its semantic accuracy on each conflict question. The 194 Gemma questions are partitioned at the empirical quartiles of this prior measure. The resulting bin sizes are 49, 48, 48, and 49. Table~\ref{tab:prior_details} supplies the values plotted in Fig.~\ref{fig:prior_corr}. The log-probability measures preference for the true answer in the question context; the pairwise margin additionally depends on the document answer's logit.
\begin{table}[H]
\centering
\caption{Gemma prior-strength quartiles with semantic answer scores (\%).}
\label{tab:prior_details}
\begin{tabular}{lrrrrr}
\toprule
Prior quartile & $n$ & Mean log-prob. & Baseline & SLB & CA \\
\midrule
Very low & 49 & -12.41 & 65.3 & 75.5 & 81.6 \\
Low & 48 & -6.27 & 72.9 & 85.4 & 89.6 \\
Medium & 48 & -2.32 & 52.1 & 81.2 & 87.5 \\
High & 49 & -0.47 & 16.3 & 30.6 & 46.9 \\
\bottomrule
\end{tabular}

\end{table}

\subsection{Mistral Prior-Strength Bins}
\label{app:mistral_prior}

\begin{table}[H]
\centering
\caption{Substring-scored Mistral prior-strength analysis. Accuracy in each prior-strength bin defined by value thresholds on the base model's log-probability of the true pretraining answer (cut points at $-10$, $-5$, and $-2$, giving the bin sizes shown in column $n$). The baseline gap between the weakest and strongest prior bins is 36 percentage points; Conflict-Aware reduces the gap to 1 point.}
\label{tab:mistral_prior}
\begin{tabular}{lcccc}
\toprule
Prior strength (bin) & n & Baseline & SLB & Conflict-Aware \\
\midrule
Very low (avg log-prob $-13.9$) & 42 & 55\% & 71\% & 76\% \\
Low (avg $-7.2$) & 86 & 59\% & 70\% & 72\% \\
Medium (avg $-3.4$) & 50 & 58\% & 78\% & 78\% \\
High (avg $-1.7$) & 16 & 19\% & 44\% & \textbf{75\%} \\
\bottomrule
\end{tabular}
\end{table}

\section{Hyperparameters, Hardware, and Latency}
\label{sec:hyper-hw}
\label{app:timing}

\begin{table}[H]
\centering
\caption{Primary evaluation settings. Other benchmarks use the budgets and subsets stated in their respective sections.}
\label{tab:hyperparams}
\begin{tabular}{lll}
\toprule
Component & Parameter & Value\\
\midrule
Generated LoRA & rank / scale & $8$ / $45.25$\\
Target modules & & MLP down projection\\
SLB & layer fraction & $25\%$ (rounded down)\\
SLB & boost & $1.75$ Gemma/Mistral; $3$ Qwen\\
CA strong branch & layer fraction & Approximately one third\\
CA strong branch & boost & $2$ Gemma/Mistral; $3$ Qwen\\
CA probe & generation budget & $20$ tokens\\
Primary answers & decoding / budget & Greedy / $64$ tokens\\
Primary scoring & & Answer-level semantic\\
Controlled SLB intervals & document bootstrap & $5{,}000$ resamples\\
Inference & dtype & bfloat16\\
\bottomrule
\end{tabular}
\end{table}

The released hypernetworks are used without retraining: Gemma-2-2B-IT at 80K steps, Mistral-7B-Instruct-v0.2 at 20K, and Qwen3-4B-Instruct-2507 at 20K. Timing uses one NVIDIA A800-SXM4-80GB. Each A/C pool contains the first query from 15 documents, measured for three rounds (45 observations per method and pool). Table~\ref{tab:latency} gives mean query latency with the generated adapter already available. Each method can generate a different answer length, so the reported time includes that decoding behavior.

\begin{table}[H]
\centering
\caption{Mean query latency in seconds on one A800, 45 measurements per cell. CA includes its 20-token probe.}
\label{tab:latency}
\begin{tabular}{llrrrr}
\toprule
Model & Pool & Baseline & SLB & Always-strong & CA \\
\midrule
Gemma-2B & A & 0.313 & 0.315 & 0.310 & 0.899 \\
Gemma-2B & C & 0.353 & 0.320 & 0.317 & 0.866 \\
Mistral-7B & A & 1.000 & 1.040 & 1.122 & 1.589 \\
Mistral-7B & C & 1.434 & 1.401 & 1.411 & 1.893 \\
Qwen-4B & A & 0.824 & 0.964 & 0.864 & 1.490 \\
Qwen-4B & C & 1.248 & 0.879 & 0.812 & 1.456 \\
\bottomrule
\end{tabular}

\end{table}

\paragraph{Reproducibility.} The accompanying code specifies generation settings, seeds, update-norm calculations, and answer scoring. The data include model responses and semantic judgments used to generate the tables and figures.

\clearpage
\flushbottom
\section*{NeurIPS Paper Checklist}

\begin{enumerate}

\item {\bf Claims}
    \item[] Question: Do the main claims made in the abstract and introduction accurately reflect the paper's contributions and scope?
    \item[] Answer: \answerYes{} 
    \item[] Justification: The abstract and introduction state the magnitude account, training-free interventions, and benchmark contribution. The main results separate boosting, matched-norm allocation, and routing effects, with their scope discussed in Section~\ref{sec:discussion}.

\item {\bf Limitations}
    \item[] Question: Does the paper discuss the limitations of the work performed by the authors?
    \item[] Answer: \answerYes{} 
    \item[] Justification: The limitations have been discussed in Section~\ref{sec:discussion}.

\item {\bf Theory assumptions and proofs}
    \item[] Question: For each theoretical result, does the paper provide the full set of assumptions and a complete (and correct) proof?
    \item[] Answer: \answerYes{} 
    \item[] Justification: The pairwise margin condition is derived in Section~\ref{sec:theory}, which states the local-response approximation and its scope. The empirical predictions are examined through prior-strength analysis, scaling interventions, and routing comparisons.

    \item {\bf Experimental result reproducibility}
    \item[] Question: Does the paper fully disclose all the information needed to reproduce the main experimental results of the paper to the extent that it affects the main claims and/or conclusions of the paper (regardless of whether the code and data are provided or not)?
    \item[] Answer: \answerYes{} 
    \item[] Justification: All information needed to reproduce the main results is disclosed in the paper and appendix, and open-source assets and materials necessary for reproducing have been properly cited. The supplement specifies scoring, answer references, matched-norm controls, and uncertainty calculations.

\item {\bf Open access to data and code}
    \item[] Question: Does the paper provide open access to the data and code, with sufficient instructions to faithfully reproduce the main experimental results, as described in supplemental material?
    \item[] Answer: \answerYes{}
    \item[] Justification: Code, data, response records, plotting scripts, and reproduction instructions are available at \url{https://github.com/KigML/OverrideGap}.

\item {\bf Experimental setting/details}
    \item[] Question: Does the paper specify all the training and test details (e.g., data splits, hyperparameters, how they were chosen, type of optimizer) necessary to understand the results?
    \item[] Answer: \answerYes{} 
    \item[] Justification: All experimental details are specified. The evaluation setup is in Section~\ref{sec:kidbench} and Appendix~\ref{app:kidbench}. A complete hyperparameter table is given in Section~\ref{sec:hyper-hw}.

\item {\bf Experiment statistical significance}
    \item[] Question: Does the paper report error bars suitably and correctly defined or other appropriate information about the statistical significance of the experiments?
    \item[] Answer: \answerYes{} 
    \item[] Justification: Paired document-bootstrap intervals accompany the matched-norm comparisons in Appendix~\ref{app:controls}. Routing comparisons report paired intervals, matched-rate random allocations, and fixed-branch ceilings in Appendix~\ref{app:ca_impl}. Label uncertainty is reported separately from sampling uncertainty.

\item {\bf Experiments compute resources}
    \item[] Question: For each experiment, does the paper provide sufficient information on the computer resources (type of compute workers, memory, time of execution) needed to reproduce the experiments?
    \item[] Answer: \answerYes{} 
    \item[] Justification: Checkpoints, dtype, hardware, generation budgets, and measured query latency are specified in Appendix~\ref{sec:hyper-hw}. All main interventions use frozen released hypernetworks; the accompanying run configurations specify timing and hardware.

\item {\bf Code of ethics}
    \item[] Question: Does the research conducted in the paper conform, in every respect, with the NeurIPS Code of Ethics \url{https://neurips.cc/public/EthicsGuidelines}?
    \item[] Answer: \answerYes{} 
    \item[] Justification: We have reviewed NeurIPS ethics guidelines and confirm that we follow all the code of ethics.


\item {\bf Broader impacts}
    \item[] Question: Does the paper discuss both potential positive societal impacts and negative societal impacts of the work performed?
    \item[] Answer: \answerYes{} 
    \item[] Justification: See Ethics paragraph in Section~\ref{sec:discussion}.

\item {\bf Safeguards}
    \item[] Question: Does the paper describe safeguards that have been put in place for responsible release of data or models that have a high risk for misuse (e.g., pre-trained language models, image generators, or scraped datasets)?
    \item[] Answer: \answerNA{} 
    \item[] Justification: The new artifact is an authored benchmark; it contains no personally identifiable information, no images, and no internet-scraped content. The proposed methods are training-free post-processing of an already-public hypernetwork and do not introduce new pretrained weights.

\item {\bf Licenses for existing assets}
    \item[] Question: Are the creators or original owners of assets (e.g., code, data, models), used in the paper, properly credited and are the license and terms of use explicitly mentioned and properly respected?
    \item[] Answer: \answerYes{} 
    \item[] Justification: Existing assets used: SQuAD~\citep{squad} (CC BY-SA 4.0), GSM8K~\citep{gsm8k} (MIT), TruthfulQA~\citep{truthfulqa} (Apache-2.0), Gemma-2-2B-IT~\citep{gemma} (Gemma Terms of Use), Mistral-7B-Instruct-v0.2~\citep{mistral7b} (Apache-2.0), Qwen3-4B-Instruct-2507~\citep{qwen} (Apache-2.0), Doc-to-LoRA pretrained hypernetwork checkpoints~\citep{doc2lora} per their original release. We confirm compliance with each license's redistribution and attribution requirements.

\item {\bf New assets}
    \item[] Question: Are new assets introduced in the paper well documented and is the documentation provided alongside the assets?
    \item[] Answer: \answerYes{} 
    \item[] Justification: The benchmark construction, answer references, intended use, scoring criteria, and limitations are documented in the paper and supplement. The repository includes the code and record-level documentation.

\item {\bf Crowdsourcing and research with human subjects}
    \item[] Question: For crowdsourcing experiments and research with human subjects, does the paper include the full text of instructions given to participants and screenshots, if applicable, as well as details about compensation (if any)?
    \item[] Answer: \answerNA{} 
    \item[] Justification: This paper does not involve crowdsourcing nor research with human subjects. KID-Bench was authored by the authors themselves (Appendix~\ref{app:kidbench}), with no external annotators or paid workers.

\item {\bf Institutional review board (IRB) approvals or equivalent for research with human subjects}
    \item[] Question: Does the paper describe potential risks incurred by study participants, whether such risks were disclosed to the subjects, and whether Institutional Review Board (IRB) approvals (or an equivalent approval/review based on the requirements of your country or institution) were obtained?
    \item[] Answer: \answerNA{} 
    \item[] Justification: This research does not involve human subjects, so IRB approval (or equivalent) is not applicable. The benchmark questions and documents in KID-Bench were authored by the authors themselves (Appendix~\ref{app:kidbench}); no participants were recruited and no human-subject data was collected.

\item {\bf Declaration of LLM usage}
    \item[] Question: Does the paper describe the usage of LLMs if it is an important, original, or non-standard component of the core methods in this research? Note that if the LLM is used only for writing, editing, or formatting purposes and does \emph{not} impact the core methodology, scientific rigor, or originality of the research, declaration is not required.
    \item[] Answer: \answerYes{}
    \item[] Justification: The CA probe and semantic answer evaluation use language models. The probe is described in Appendix~\ref{app:ca_impl}, and the evaluation rubric, judge, and treatment of uncertain labels are specified in Appendix~\ref{app:scoring}.

\end{enumerate}

\end{document}